\documentclass[conference]{IEEEtran}
\IEEEoverridecommandlockouts
\usepackage{cite}
\usepackage{amsmath,amssymb,amsfonts}
\usepackage{algorithmic}
\usepackage{graphicx}
\usepackage{textcomp}
\usepackage{xcolor}
\usepackage {CJKutf8}
\def\BibTeX{{\rm B\kern-.05em{\sc i\kern-.025em b}\kern-.08em
    T\kern-.1667em\lower.7ex\hbox{E}\kern-.125emX}}
\begin{document}

\title{DiReDi: Distillation and Reverse Distillation for AIoT Applications}

\author{CHEN SUN 1, SENIOR MEMBER, IEEE, QIANG TONG 1, WENSHUANG YANG 2,\\WENQI ZHANG 2,  MEMBER, IEEE, \\1 SONY China Research Laboratories, Beijing 100027, China, \\2 Nanyang Technological University, 639798, Singapore}

\maketitle

\begin{abstract}
Artificial Intelligence $\&$ Internet of Things (AIoT) have been widely utilized in various application scenarios. Typically, the significant efficiency can be achieved by deploying different edge-AI models in various real-world scenarios while a few large models manage those edge-AI models remotely from cloud servers. However, customizing edge-AI models for each user's specific application or extending current models to new application scenarios remains a challenge. Inappropriate local training or fine-tuning of edge-AI models by users can lead to model malfunction, potentially resulting in legal issues for the manufacturer. To address aforementioned issues, this paper proposes an innovative framework called ''DiReD'', which involves knowledge \textbf{Di}stillation $\&$ \textbf{Re}verse \textbf{Di}stillation. In the initial step, an edge-AI model is trained with presumed data and a KD process using the cloud AI model in the upper management cloud server. This edge-AI model is then dispatched to edge-AI devices solely for inference in the user's application scenario. When the user needs to update the edge-AI model to better fit the actual scenario, the reverse distillation (RD) process is employed to extract the knowledge — the difference between user preferences and the manufacturer's presumptions from the edge-AI model using the user's exclusive data. Only the extracted knowledge is reported back to the upper management cloud server to update the cloud AI model, thus protecting user privacy by not using any exclusive data. The updated cloud AI can then update the edge-AI model with the extended knowledge. Simulation results demonstrate that the proposed ``DiReDi" framework allows the manufacturer to update the user model by learning new knowledge from the user's actual scenario with private data. The initial redundant knowledge is reduced since the retraining emphasizes user private data. Furthermore, this model update approach via cloud allows manufacture to check model updates ensuring that all models are managed safely and effectively.
\end{abstract}

\begin{IEEEkeywords}
Artificial Intelligence $\&$ Internet of Things (AIoT), Knowledge Distillation (KD), object detection, edge-AI
\end{IEEEkeywords}


\section{INTRODUCTION}
\begin{figure*}[t]
    \centering
    \includegraphics[width=\linewidth]{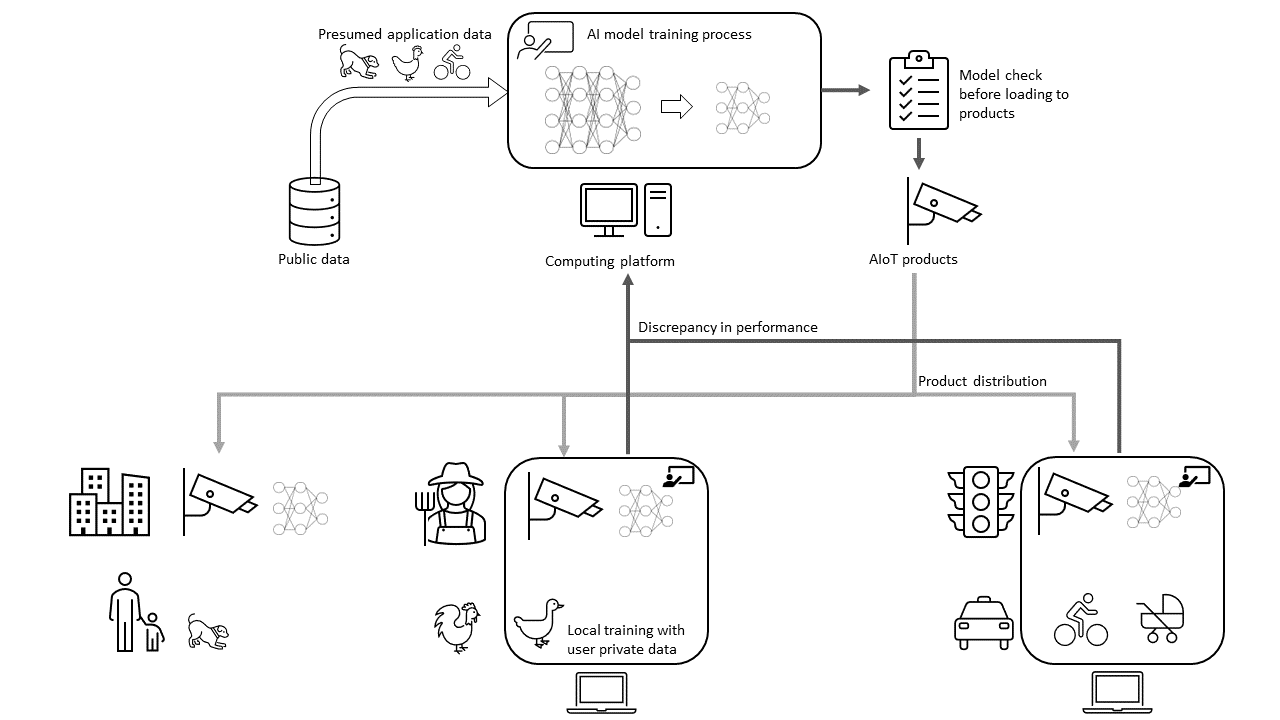}
    \caption{Scenario of AIoT product applications and model training.}
    \label{Fig.scenario}
\end{figure*}

\IEEEPARstart{W}{ith} the widespread adoption of the Internet of Things (IoT) communication networks and Artificial Intelligence (AI) technologies, Artificial Intelligence $\&$ Internet of Things (AIoT) applications have become prevail in human daily life \cite{AIoT,ZhangAIOT21}. Especially, edge-AI based AIoT systems, as shown in Fig.~\ref{Fig.scenario}, have become a trend due to their great potential for low profile, minimal power supply, and closer proximity to users compared to traditional systems that rely on normal sensors on the user's side and AI models deployed solely on remote cloud servers. In short, edge-AI based AIoT systems offer low power consumption due to their portable sizes, along with minimal bandwidth costs resulting from the elimination of continuous transmission of large scale data such as high-quality images to cloud servers. This renders edge devices ideal for meeting the portability and low power consumption needs of real-world scenarios, providing end-users with cost-effective and easily deployable systems. Recently, There are several applications examples that show the advantages of edge-AI based AIoT systems. For example, an AI model can be deployed on devices for antenna radiation direction control as in \cite{ZhuNetwork20,Sun-24-OJCOM}. In agricultural applications, the smart camera is used to detect potential sickness in chickens \cite{Tong-23}. Applying AI on cars help obstacle detection in autonomous driving\cite{sun22,Pasricha23}. In smart city applications, the smart cameras are deployed to detect pedestrians running into the street \cite{Solmaz-19,TaoVTC24}. 

Even though, edge-AI based AIoT systems offer the advantage of proximity to users, upgrading the edge-AI models to better fit each individual and specific application scenario remains a challenge due to some issues such as data transfer, legal concerns, etc. As an example shown in Fig.~\ref{Fig.scenario}, distributors such as companies that manufacture AIoT products, including edge-AI models equipped with edge-AI enabled devices, typically train and deploy these models for specific application scenarios such as city security monitoring, smart livestock farming, and traffic surveillance. However, the model training usually relies on general presumed data, which cannot perfectly cover all the varied and specific application scenarios. This leads to an accuracy gap in the edge-AI model for real-world scenarios due to the differences between the training data presumed by the AIoT company and the data from actual scenarios. For example, as shown in Fig.~\ref{Fig.scenario}, the AIoT product with an edge-AI model is originally designed to detect only chickens (layers and broilers). However, in actual scenarios, it might be necessary to detect other livestock, such as ducks and cows, etc. This necessitates an efficient method to extend the application scenarios for the edge-AI model. Furthermore, allowing users to conduct local model training or fine-tuning, or permitting AIoT companies to retrain the AI models using large-scale data from actual scenarios, is rarely a good solution. This is due to numerous issues such as privacy concerns regarding the received data and accountability issues for AI models trained locally by users. For instance, local training on the user side for traffic surveillance might enhance the accuracy of detecting bicycles and baby carriages, thus better fitting the actual scenario. However, this could ``damage" the original knowledge for pedestrian detection, potentially causing serious consequences and leading to confused accountability between the AIoT company and users. Hence, a better approach is to update the edge-AI model of the AIoT company by using the ``knowledge" rather than the actual data from the user side, and then dispatch the upgraded edge-AI model back to the users to better fit the actual application scenarios. This method can resolve the mentioned issues related to data privacy and accountability.  

In this article, we propose a \textbf{Di}stillation and \textbf{Re}serve \textbf{Di}stillation (DiReDi) framework to update the model of the AIoT company and the corresponding edge-AI model on the user side. This framework updates the ``knowledge," including features and neuron weights, based on the actual application scenario instead of transferring raw data from the user to the AIoT company. Therefore, this approach is capable to avoid the issues related to data privacy and accountability. In detail, this approach is based on knowledge distillation (KD) techniques. First, we consider the edge-AI model dispatched from the AIoT company to the actual scenario as being trained on presumed data using the KD process. This means the AIoT company owns a large AI model as a teacher model and a lightweight edge-AI model as a student model. By using presumed scenario data and KD, the edge-AI model achieves high accuracy with a light-weighted model size. This lightweight edge-AI model is then dispatched with the edge-AI device such as edge-AI enabled CMOS sensor, Nvidia Jetson Nano, etc. to the actual application scenario solely for inference purposes. It is worth noting that KD from a larger AI model is an effective way to train an edge-AI model for better accuracy compared to directly training the edge-AI model, as demonstrated by the experimental results of \cite{KD-tong}. To enhance the accuracy of the dispatched edge-AI model on the user side using data from actual scenarios rather than presumed data, which may not fully cover the actual scenario, the model difference is simulated using two relatively larger models defined as tutor models based on ``reverse distillation (KD)". In this context, KD means the teacher model is a smaller-sized model (edge-AI model) and the student model is a larger-sized model (tutor model), in contrast to traditional KD process, which utilizes a larger-sized model as the teacher and a smaller-sized model as the student. Furthermore, one of the two tutor models is used to emulate the behavior of the edge-AI model, while the other tutor model emulates the accuracy expected by the user in the actual scenario. The difference between the two tutor models, which denotes the discrepancy in knowledge, can be uploaded from the actual scenario to the AIoT company via the internet or other communication channels. This discrepancy can be utilized to update the AIoT company's original teacher model without using raw data, thereby avoiding privacy issues. After the teacher model is updated by the AIoT company, the student model can also be upgraded using the corrected teacher model through a KD process and then redeployed to the actual application scenario. This approach avoids legal and accountability issues associated with edge-AI model fine-tuning in local scenarios by users. To the best of our knowledge, this is the first time the actual scenario data of the student model has been used to provide extended knowledge to the teacher model, specifically enriching the teacher model's generalization ability.
 
A simulator is established to train and test the performance of the proposed DiReDi framework. In this paper, we consider object detection as the application task and all AI models utilized in this paper are detectors. We select Fully Convolutional One-Stage (FCOS) detector \cite{FCOS2019} with ResNet50 as the teacher model, FCOS-Lite detector \cite{KD-tong} with MobileNet as student model. The focal and global KD (FGD) \cite{kd17} is selected as the basic KD process. simulation carried out to demonstrate that the proposed DiReDi framework allows the manufacturer to update the user model by learning the new extended knowledge while the unused knowledge will be discarded.

The rest of this article is organized as follows. Section demonstrates the preliminaries. The proposed DiReDi framework is reviews in Section III. Section IV elaborates the experiment. Section V shows the simulation results followed by the concludes in Section VI.

\section{PRELIMINARIES}\label{sec2}
An AIoT equipment has both energy consumption and computation power constrains. The AI models at AIoT devices shall provide satisfactory inference performance with limited size. Therefore, selecting an appropriate method to compress the model is crucial to meet deployment requirements, including the computational power and response time constraints.

\subsection{AIoT $\&$ OBJECTIVE DETECTION}
In recent years, there has been notable attention directed towards the rapid advancements in AIoT (AI and IoT) technologies, such as smart agriculture field \cite{intro1, intro3}. With the evolution of AIoT, edge computing has ushered in a new era of practical applications in recent AIoT fields. Unlike high-performance computing equipment or cloud servers, edge computing devices feature an open platform that integrates network, computing, storage, and application core capabilities, delivering services closest to the source of data or events. Moreover, these devices offer low power consumption due to their compact size and weight, along with minimal bandwidth costs resulting from the elimination of continuous transmission of high-quality images to cloud servers. This renders edge devices ideal for meeting the portability and low power consumption needs of real-world scenarios, providing end-users with cost-effective and easily deployable systems. However, the constrained storage capacity and computing capabilities of edge devices pose challenges for deploying high-performance AI models when compared to high-performance computing equipment or cloud servers. Hence, it is crucial to explore effective AI models that offer compact size, minimal computing power requirements, and yet maintain good performance.

Recently, many deep learning-based object detection technologies showed their, such as fast region-based convolutional neural network (fast-RCNN) \cite{intro7}, feature fusion single shot (FSSD) \cite{intro8} and you only look once (YOLO) \cite{intro9} series, especially, current YOLOv5 \cite{ex1} shows high performance and well used in a lot of application scenarios. Compared with YOLO based method which are anchor based methods, FCOS \cite{FCOS2019} is a proposal-free, anchor-box-free, single-stage object detection model. By eliminating the predefined set of anchor boxes, FCOS circumvents the computations and hyper-parameters for anchor boxes, which often significantly influence final detection performance. Furthermore, FCOS  exhibits a simple structural design yet delivers robust performance in object detection tasks. Its architecture enables easy adjustment of complexity to suit a variety of detection tasks. Hence, to adapt the FCOS detector for edge devices, we introduce a FCOS-Lite detector, which is a streamlined version optimized for lightweight processing.

\subsection{KNOWLEDGE DISTILLATION}\label{sec2.3}
In general, there are typically two kinds of KD processes: logits mimicking based and feature imitation based\cite{wang2021knowledge,gou2021knowledge}. The logits mimicking distillation is first used in the distillation of classification tasks\cite{hinton2015distilling}, which aims to minimize the discrepancy between the  output layer of the student model and the teacher model. Due to the clear physical interpretation of logits: the predicted probabilities of each object category, it is easier to implement and optimize. The loss function can be represented as \cite{gou2021knowledge}
\begin{equation}
\begin{split}
& L_{distill}  = \\
&\quad l_{logits}\left(Softmax\left(z_{\text {T }}(x), t\right), Softmax\left(z_{\text {S}}(x), t\right)\right)\\
&Softmax\left(z_{i}, t\right)=\frac{\exp \left(z_{i} / t\right)}{\displaystyle \sum_{j=1}^{K}\exp \left(z_{j} / t\right)}
\end{split},
\end{equation}
where $x$ is the input feature. Here, $l_{logits}(\cdot)$ indicates the loss function to calculate the difference of logits between the teacher model and student model. $z_{\text {T}}$ and $z_{\text {S}}$ are the output logits of the teacher model and student model, respectively. $t$ is the hyper-parameters to adjust learning effect.

For the feature imitation distillation, it can pay more attention to distill and transfer richer abstract knowledge by utilizing high-dimensional features from the hidden layers of the teacher model to guide the student model \cite{2014arXiv1412.6550R}. The loss function can be represented as \cite{gou2021knowledge}: 
\begin{equation}
% \begin{split}
L_{fea}^{distill}={l}_{fea}(\mathcal R(F_{T}(x)), \mathcal R(f_{S}(x)))
% \end{split},
\end{equation}
where $l_{fea}(\cdot,\cdot)$ indicates the loss function to calculate the difference between the teacher model and student model feature maps. $\mathcal R(\cdot)$ denotes reshape operation. $F_{T}$ and $f_{S}$ are the output feature maps of the teacher model and student model, respectively, where the capital letter $F$ indicates the teacher is a large model.

Although there is still no consensus on the superiority of these two methods\cite{FCOS2019,shu2021channel,cao2022pkd,zheng2022localization}, the feature imitation distillation has demonstrated significant advantages in the task of object detection by design more rigorous distillation rules and more comprehensive distillation path, likely because it forces student models to process and understand whole features rather than predicted logits on a deeper level. Considering the diversity of actual user scenarios, which leads to the richness and uncertainty of object categories, the logits mimicking distillation may face issues such as information loss and reduced generalization ability. Thus, in our study, the feature imitation distillation is employed. Recently, there has been a growing interest in applying KD processes to detectors \cite{kd12,kd13,kd14,kd15, kd16}. Especially, the method proposed in \cite{kd17} called ``FGD" employs focal distillation and global distillation to encourage the student network to learn the critical pixels, channels, and pixel relations from the teacher network. This approach provides an efficient and accurate approach to distill models since various and fulfilled features are extracted and utilized as knowledge for distillation and only feature maps of model's neck are utilized to extract all features. To better explain DiReDi based on FGD, a brief review of FGD is given. FGD relies on three components to jointly facilitate knowledge distill and transfer: the original loss of detector, the focal distillation loss, and the global distillation loss.

As shown in Fig.~\ref{Fig.DiReDi}(b), both focal distillation and global distillation are achieved through the computation of focal and global distillation losses, which are calculated from the Feature Pyramid Networks (FPN) \cite{kl20}) of both the neck of teacher and student models. In focal distillation, ground truth bounding boxes are utilized to generate a binary mask $M$, scale mask $S$ for segregating the background and foreground within the feature map. Next, spatial and channel attention masks, denoted as $A^{s}$ and $A^{c}$ respectively, are calculated from teacher model based on attention mechanisms \cite{kl18,kl19}. These masks from the teacher model are then utilized to guide the student model in the focal distill loss as
\begin{equation}
\label{eq11}
\begin{aligned}
L_{focal} &=\sigma \sum_{k=1}^{C}\sum_{i=1}^{H}\sum_{j=1}^{W}M_{i,j}S_{i,j}A_{i,j}^{s}A_{k}^{c}(F_{k,i,j}^{T}-f_{k,i,j}^{S})^{2} \\ &+\beta\sum_{k=1}^{C}\sum_{i=1}^{H}\sum_{j=1}^{W}\hat{M}_{i,j}\hat{S}_{i,j}A_{i,j}^{s}A_{k}^{c}(F_{k,i,j}^{T}-f_{k,i,j}^{S})^{2} \\ &+\gamma(L_{1}(A_{T}^{s},A_{S}^{s})+L_{1}(A_{T}^{c},A_{S}^{c})) ,
\end{aligned}
\end{equation}
where $\sigma$, $\beta$ and $\gamma$ are hyper-parameters to balance the loss contributions between foreground, background and regularization respectively.  $F^{T}$ and $f^{S}$ denote the feature maps of the teacher detector and student detector, respectively. Index $k$, $i$ and $j$ denote the channel, height and width of feature maps, respectively. $\hat{M}$ and $\hat{S}$ represent the inverse binary mask and inverse scale mask to preserve the background within the feature map, respectively, while $L_{1}(\cdot)$ denote $L_{1}$ loss.

On the other hand, global distillation loss is utilized to capture the long-range dependencies within a single image, which can be formulated as
\begin{equation}
L_{global} = \lambda \sum(G(F_{T})-G(F_{S}))^{2} \label{eq12},
\end{equation}
with
\begin{equation}
G(F) = F+W_{2}(ReLU(LN(W_{1}(\sum_{j=1}^{N_{p}}\frac{e^{W_{k}F_{j}}}{\sum_{m=1}^{N_{p}} e^{W_{k}F_{m}}}F_{j}  )))) \label{eq13},
\end{equation}
where $\lambda$ denote a hyper-parameter, $W_{k}(\cdot)$, $W_{1}(\cdot)$, $W_{2}(\cdot)$, $ReLU(\cdot)$ and $LN(\cdot)$ represent the outputs of convolutional layers $W_{k}$, $W_{1}$, $W_{2}$, ReLU, and layer normalization, respectively. $N_{p}$ denote the number of pixels in the feature. 

The final expression for the FGD distillation loss function can be summarized as 
\begin{equation}
l_{fea}=L_{\text {focal}}+L_{\text {global}}.\label{eq41}
\end{equation}

\section{THE PROPOSED DiReDi FRAMEWORK}
\begin{figure*}
    \centering
    \includegraphics[width=0.9\linewidth]{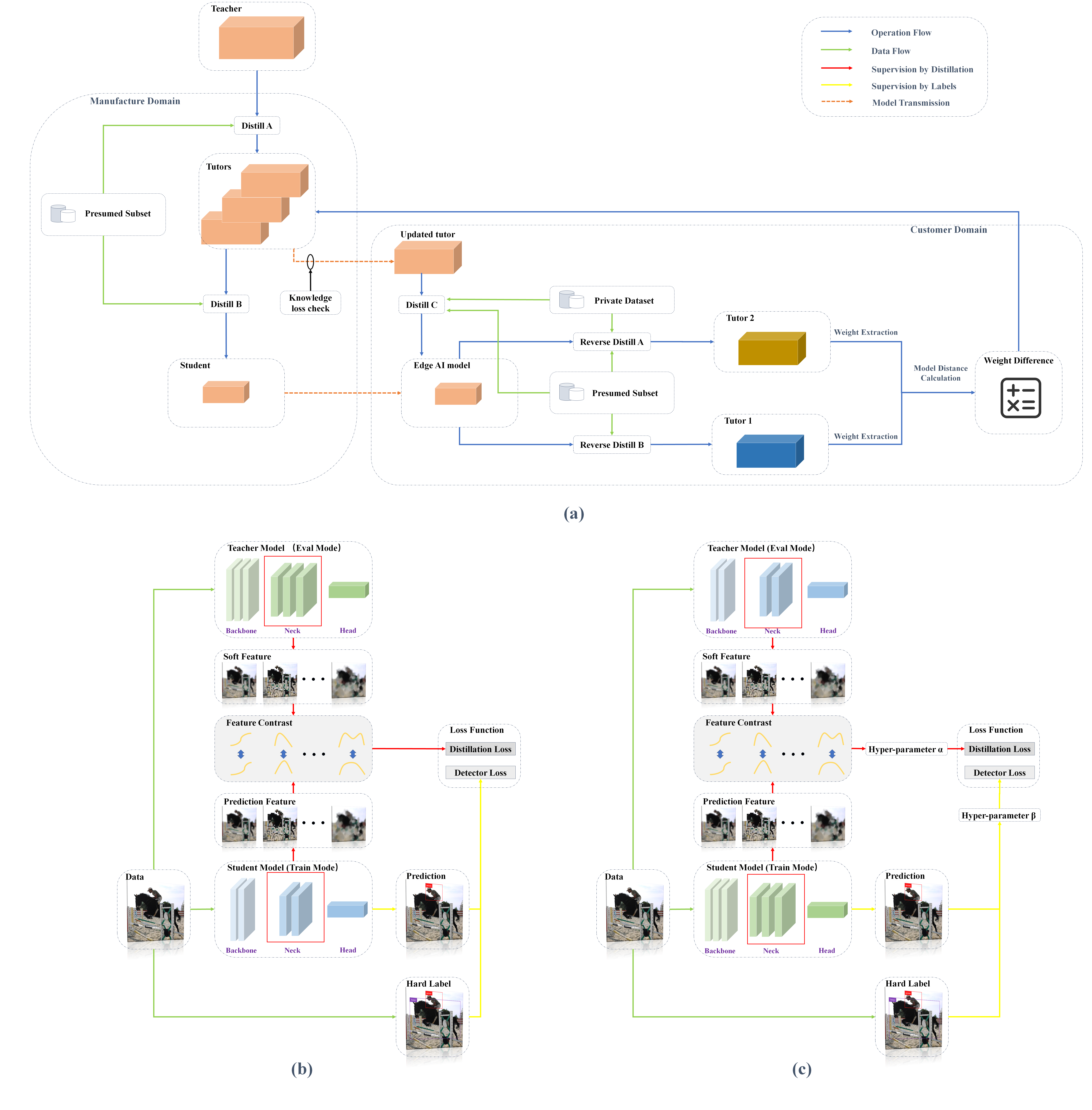}
    \caption{(a) Distillation and reverse distillation (DiReDi) for AIoT applications. (b) Knowledge distillation (KD) process. (c) Reverse distillation (RD) process.}
    \label{Fig.DiReDi}
\end{figure*}
In this section, we give the detail of the novel DiReDi framework for AIoT applications, which is shown in Fig.~\ref{Fig.DiReDi}(a). We assume that the manufacturing organizations such as an AIoT company creates edge-AI model shown as a "student" for various applications based on pre-collected data and KD process from large AI model that we call it tutor model. The tutor model is a relatively larger model made for specific application scenario. It is worth noting that the tutor model may be generated from another much larger AI model as a "teacher", which is managed by the AIoT company. Basically, there is a KD process to generate edge-AI model from tutor model in manufacture domain by using presumed data, then there is a RD process to extract the knowledge updated by private data in the customer's domain by using two tutor models. Finally, the updated knowledge from customer domain can be used for updating the original tutor model in manufacture domain without updating any real private data to avoid private issues. After the  original tutor model updated, the edge-AI model can be updated too based on KD process and the updated tutor model.       
% It is worth noting that all techniques operate on the neck and head parts of the model rather than the backbone, demonstrating broad applicability. /I think we need to write some reason here/

\subsection{DISTILLATION PROCESS}
The manufacturing domain defines the business scope of an AIoT company specializing in producing AIoT devices, such as image sensors\cite{iot2} combined with AI functions for various applications. The AIoT company can create an application-specific model, typically of intermediate size, derived from a substantial model hosted on the cloud and trained using extensive datasets. The large model's capacity allows it to achieve high accuracy on diverse and large scale data, such as a comprehensive collection of images. For instance, if the company is producing edge-AI enabled AIoT equipment for animal recognition, it can train a intermediate-size model, referred to as a ``tutor model," through KD using public data relevant to the application. Additionally, a smaller model can be further distilled from the tutor model, which can then be deployed on the AIoT equipment for data processing. As shown in Fig.~\ref{Fig.scenario}, this AIoT company generates three tutor models for three application scenarios: smart city, smart agriculture, and intelligent transportation. For each scenario, the corresponding edge-AI model is further created from the tutor model via the KD process. The detailed distillation procedure is shown in Fig.~\ref{Fig.DiReDi}(b). In this step, we employ FGD method described in subsection~\ref{sec2}.\ref{sec2.3} as the KD process. Therefore, the ``Distillation Loss" shown in Fig.~\ref{Fig.DiReDi}(b) is as described in Eqs.~(\ref{eq11})-(\ref{eq41}). The teacher  and student models of KD process are the large tutor model and the edge-AI model, respectively. 

\subsection{REVERSE DISTILLATION PROCESS}
In the customer domain, the student model of the KD process, which is the edge-AI model, is deployed on AIoT equipment on the users' (edge) side. This kind of edge-AI-enabled AIoT technology has several advantages for actual application scenarios. However, the data encountered in actual application scenarios may significantly differ from the presumed data used to generate the tutor model and the edge-AI model in the manufacturing domain. Compared to the presumed data, this kind of data from actual scenarios may include information on conditions different from those in the presumed data, as well as information from corner cases, etc. This difference may cause accuracy decreases in actual scenarios. For instance, an AIoT company might produce a smart camera with an edge-AI model designed to detect most breeds of chickens, but a customer might use it to detect a rare breed of chicken. In this case, the data related to the customer's rare breed of chickens becomes the customer's private data, and the AI technology may fail to work properly. In some scenarios, such as intelligent traffic, the accuracy degradation caused by corner cases may be crucial for human safety. Therefore, it is necessary to "recover" the accuracy of the edge-AI model to better fit the actual scenarios.   

Assuming the customer's AIoT devices can connect to additional computational resources such as local PCs and local servers for managing all AIoT devices, edge data, and local AI models, we enable the customer to retrain the same tutor model locally with assistance and the presumed data provided by the AIoT company. Instead of retraining the small model directly, since the compact model size and limited memory capacity of edge-AI models deployed on the AIoT devices only allow for the inference process, rather than the more memory-consuming training process. Hence, to improve the accuracy of the edge-AI model, we propose the RD process. In this process, we firstly employ the edge-AI model as the teacher model to train tutor models (student models), as illustrated in Fig.~\ref{Fig.DiReDi}(a) and (c). In Fig.~\ref{Fig.DiReDi}(a), "Tutor 1" is trained with the presumed data through the RD process to emulate the behavior of the edge-AI model as envisioned by the AIoT company. In contrast, "Tutor 2" is trained with the customer's additional private data captured in actual scenarios through RD, reflecting the expected behavior in the customer's application scenario. The differences between these two tutor models, "Tutor 1" and "Tutor 2," highlight the knowledge gap between the presumed data and the actual scenario data. The RD process can be described as, 
% Larger models involve more weights \textbf{\textit{W}} during the process of training and inference. Therefore, when receiving a limited number of input features, it can largely approximate the output of the smaller model due to the presence of redundant neurons. In our system, the above process is achieved by using knowledge distillation in reverse, where the smaller model serves as the teacher model and the larger model serves as the student model. During the reverse distillation process, the teacher model is always set to the eval mode to output stable and consistent features, and the student model is set to the train mode and guided by the teacher model in iterations. The reverse distillation can be defined as follows:
\begin{equation}
L_{fea}^{ReDi}=l_{fea}(\mathcal R(f_{T}(x)), \mathcal R(F_{S}(x)))
\end{equation}
where $l_{fea}(\cdot,\cdot)$ denotes the RD loss which is calculated by feature maps from neck part of both teacher and student model, $R(\cdot)$ denotes reshape operation, $f_{T}$ denotes the output feature maps of the small model as a teacher in forward only evaluation mode, $F_{s}$ denotes the output feature maps of the tutor model as a student.

When the edge-AI model, with its smaller size, serves as the teacher and the tutor model, with its larger size, serves as the student in the RD process, the tutor model (student) can easily approximate the outputs and performance of the teacher (small model) due to the greater number of weights and neurons in the larger student model. Furthermore, in order to simulates the behavior of the teacher (edge-AI model), rather than being overly influenced by the hard label of the dataset, the proportion of knowledge sourced of the student model is constrained through the total loss function, which can be defined as 
\begin{equation}
L_{total}=\alpha L_{fea}^{ReDi} + \beta L_{detect},
\label{eq.Loss_revdistill}
\end{equation}
where $\alpha$ and $\beta$ are the the hyper-parameters to balance the source of knowledge.  

As shown in Fig.~\ref{Fig.DiReDi}(a) the RD processes for tutor~1 and tutor 2 have different training data. For the RD process A, the input data for model training contains the user private data. To emulate the behaviour of edge-AI model in this practical situation, we can set $\beta$ close to 0. This indicates that the tutor model completely trusts the edge-AI model as a teacher. 

\subsection{WEIGHT SUBSTITUTION}
Inspired by the ideology of transfer learning for object detection applications \cite{pan2009survey}, we extend the detection capability of the original tutor model by updating its neck and head parts using the knowledge gap, rather than retraining the original tutor model with users' private data, which could lead to issues such as legal complications.The knowledge gap in the user's AIoT device, implemented in actual scenarios, arises from the discrepancy between the presumed data and the users' private data used during the RD process. By using the same structure for tutor models—one emulating the behavior of the edge-AI model trained with presumed data at the manufacturer domain and the other trained with users' private data in actual scenarios—we can transfer the knowledge gap extracted from these two tutor models back to the AIoT company in the manufacturing domain. This process, as shown in Fig.~\ref{Fig.DiReDi}(a), allows for updating the tutor model in the manufacturing domain without compromising user privacy. 

The weights of the head and neck parts are extracted from the tutor 1 and tutor 2, which can be respectively represented as $W_{t1}$ and $W_{t2}$. Since the tutor 1 is trained based on a presumed data and tutor 2 is trained based on both a presumed data and user's private data, the following calculation can be performed to obtain the difference between the model weight,
\begin{equation}
\Delta W =\gamma W_{t2}-W_{t1},
\label{sub}
\end{equation}
where $\gamma$ is the the hyper-parameters to determine the calculation method.

We employ the weights of the neck and head parts instead of those of the backbone for several reasons: (1) The backbone network involves extensive nonlinear processes, making it difficult to effectively express knowledge discrepancies through parameter subtraction alone. In contrast, the neck and head parts are simpler and more linear. (2) The weight size of the neck and head parts is smaller than that of the backbone, so transferring these weights saves bandwidth and resources. It is worth noting that in our simulations, we also evaluated performance by incorporating updates from the backbone. The results were inferior to those obtained without updating the backbone, thereby justifying our chosen approach.

The updated model reload the weight of the neck and head parts in a similar manner after receiving the different weight from customer domain, thereby obtaining the capacity to detect new object categories without extra training. The updating method is as
\begin{equation}
\Hat{W} = W_{origin}+\delta  \Delta W,
\end{equation}
where $\Hat{W}$ and $W_{origin}$ denote the updated weights and original weights of the tutor model, respectively. And, $\delta$ is the hyper-parameter to determine the contribution from the update.

% A key direction in the field of transfer learning is to enhance the performance of model $f_{T}(\cdot)$ in the target domain $D_{T}=\left\{X_{T},f_{T}(X)\right\}$ by utilizing the data $X_{S}$ from the source domain $D_{S}=\left\{X_{S}, f_{S}(X)\right\}$, when given the same learning task \textbf{\textit{T}}\cite{pan2009survey}. Therefore, it is a common way to employ a classifier with high accuracy and a feature extractor with strong generalization capabilities. In our study, we have adopted a similar approach to expand the recognizable range of the model without training. Furthermore, distinct from traditional methods involving transferring the weight of the neck and head parts, which pose the risk of parameter leakage and data leakage, the updated model will be provided with a set of relative parameter values. 

\subsection{RE-DISTILLATION}
Once the tutor model is updated on the manufacturer's side, such as by the AIoT company, the company can verify if the update causes any unintended performance degradation. For instance, if the company has trained a original tutor model for pedestrian detection in intelligent transportation systems (ITS) applications, RD on the customer's side might degrade this knowledge, potentially leading to missed detection in ITS applications and subsequent customer complaints.

If the update passes verification, the company will then update the tutor model in the customer domain. By applying distillation again with this updated tutor model, the edge-AI model can also be updated with the additional knowledge and then dispatched back to the customer's domain. It is important to note that the student model is not directly retrained with the user's private data. In the following section, we will demonstrate that the performance achieved through re-distillation with the updated tutor model is superior to that of direct training.

\section{EXPERIMENTS}
\subsection{KNOWLEDGE DISTILLATION}
The models employed in all parts are variants of the object detection model FCOS \cite{FCOS2019}, comprised of three main components: backbone, neck, and head \cite{o2015introduction}. There are three types of AI models:

a) Very Large Model: A huge model located on a cloud server and managed by the AIoT company, used for generating various specific AI models for different application scenarios.
b) Tutor Models: Generated from the very large model for specific scenarios.
c) Edge-AI Model: Deployed on the AIoT equipment in actual application scenarios.

In particular, we employ ResNet101 and ResNet50 \cite{he2016deep} as the backbones of the very large model and tutor model, respectively, while MobileNetv2 \cite{sandler2018mobilenetv2} is used as the backbone of the edge-AI model. The FPN \cite{lin2017feature} is employed as the neck part for feature fusion.

It should be noted that our models have undergone lightweight processing \cite{FCOS2019}. The neck part selectively integrates feature maps with strides of 8, 16, and 32 (for ResNet101 and ResNet50, the feature maps named C2, C3, and C4. edge-AI model with MobileNetv2 backbone, the model architecture can be found in \cite{KD-tong}. In all models, the head parts only receive feature maps from the three outputs of the corresponding neck parts.

FGD is employed to distill tutor models from the very large model we call it teacher model as shown in Fig.~\ref{Fig.DiReDi}(a), and employed to distill edge-AI models from tutors. Firstly, at the central cloud, the tutor model is distilled from teacher with training data of object from those categories that the user is more concerned about. Secondly, the tutor model is used to distill knowledge in the same manner to generate the edge-AI model. These processes achieves a significant compression. The entire distillation process relies solely on publicly available datasets that have been made accessible to users in advance.

\subsection{REVERSE KNOWLEDGE DISTILLATION}
The RD process employs FGD but with edge-AI model as teacher and the tutor model as student. The detection results of the edge-AI model is emulated by the tutor 1 model as trained by presumed data. By using both publicly available datasets and the private dataset, the tutor 2 model can not only learn from the edge-AI model, but also learn new knowledge. In this process, we assume that the user can label the private dataset, thus the detector can correct nonrecognition to achieve a wider recognizable range, through the loss $L_{detect}$ in Eq.~(\ref{eq.Loss_revdistill}).

\subsection{WEIGHT SUBSTITUTION AND UPDATE}

The neck and head parts of the tutor1 and tutor2 models are extracted separately, and the weights difference of the models is obtained by performing a linear subtraction at a 1:1 ratio according to the method described in Eq.(\ref{sub}). Similarly, the weights difference is provided to update original tutor model at the same 1:1 ratio, and updated tutor model gain the ability to recognize new object categories. Finally, through an additionally fine-tuning distillation from the tutor model to the edge-AI model with both public and private dataset.

\subsection{DATASET}
Our validation is based on the Pascal VOC 2007 and 2012 dataset \cite{everingham2010pascal}. Specifically, 10 object categories with a large number of images are randomly selected to train the very large model with ResNet101 backbone. Subsequently, 5 out of 10 object categories representing the data, presumably to be encountered by potential users, are utilized to distill knowledge to the tutor model with ResNet50 backbone and the student model with MobileNetV2 backbone. The dataset can also be made available to users, which will be used at the user side for RD. When the user wants to change one or some categories in this presumed dataset with the user's private dataset, the presumed dataset can be directly processed by the user, such as replace some categories of presumed dataset with categories of user's private data. 
% If the user lose interest in a specific object categories later on, simply remove that class from the dataset before proceeding. Additionally, the user can put new data as private dataset and use that with the public dataset for the reverse distillation of Tutor 2 Model. 
This kind of data process leads to the change of knowledge during the RD. In our experiments, we separately validated the system’s capabilities for knowledge acquisition and forgetting. 
1). Experiment~1 is conducted to confirm the system’s ability to recognize an additional category of objects in the user’s actual scenario. 2). Experiment~2 is conducted to test the system’s capacity to forget one category of objects in the user’s actual scenario while adding a new category of objects. The details of all datasets are presented in the Table~\ref{tab:t1} and Table~\ref{tab:t2}.

\begin{table*}[]
\centering 
\caption{The Objects Categories Contained in Datasets for Experiment 1}
\label{tab:t1} 

\begin{tabular}{lllllll}
\hline
CLASS     & Teacher Model & Distillation A & Distillation B & Reverse Distillation A & Reverse Distillation B & Update Distillation \\ \hline
aeroplane & \multicolumn{1}{c}{\checkmark}               & \multicolumn{1}{c}{\checkmark}                & \multicolumn{1}{c}{\checkmark}                &                        \multicolumn{1}{c}{\checkmark}&                        \multicolumn{1}{c}{\checkmark}&                     \multicolumn{1}{c}{\checkmark}\\
bird      & \multicolumn{1}{c}{\checkmark}               &                &                &                        &                        &                     \\
bus       & \multicolumn{1}{c}{\checkmark}               & \multicolumn{1}{c}{\checkmark}                 & \multicolumn{1}{c}{\checkmark}                &                        \multicolumn{1}{c}{\checkmark}&                        \multicolumn{1}{c}{\checkmark}&                     \multicolumn{1}{c}{\checkmark}\\
car       & \multicolumn{1}{c}{\checkmark}               &                &                &                        &                        &                     \\
 cat& & & & \multicolumn{1}{c}{}& \multicolumn{1}{c}{\checkmark}&\multicolumn{1}{c}{\checkmark}\\
cow       & \multicolumn{1}{c}{\checkmark}               &                &                &                        &                        &                     \\
 dog& & & & & &\\
horse     & \multicolumn{1}{c}{\checkmark}               & \multicolumn{1}{c}{\checkmark}                &                \multicolumn{1}{c}{\checkmark}&                        \multicolumn{1}{c}{\checkmark}&                        \multicolumn{1}{c}{\checkmark}&                     \multicolumn{1}{c}{\checkmark}\\
motorbike & \multicolumn{1}{c}{\checkmark}               & \multicolumn{1}{c}{\checkmark}                &                \multicolumn{1}{c}{\checkmark}&                        \multicolumn{1}{c}{\checkmark}&                        \multicolumn{1}{c}{\checkmark}&                     \multicolumn{1}{c}{\checkmark}\\
person    & \multicolumn{1}{c}{\checkmark}               & \multicolumn{1}{c}{\checkmark}                &                \multicolumn{1}{c}{\checkmark}&                        \multicolumn{1}{c}{\checkmark}&                        \multicolumn{1}{c}{\checkmark}&                     \multicolumn{1}{c}{\checkmark}\\
sheep     & \multicolumn{1}{c}{\checkmark}               &                &                &                        &                        &                     \\
tvmonitor & \multicolumn{1}{c}{\checkmark}               &                &                &                        &                        &                     \\ \hline
\end{tabular}
\end{table*}
\begin{table*}[]
\centering 
\caption{The Objects Categories Contained in Datasets for Experiment 2}
\label{tab:t2} 

\begin{tabular}{lllllll}
\hline
CLASS     & Teacher Model & Distillation A & Distillation B & Reverse Distillation A & Reverse Distillation B & Update Distillation \\ \hline
aeroplane & \multicolumn{1}{c}{\checkmark}               & \multicolumn{1}{c}{\checkmark}                & \multicolumn{1}{c}{\checkmark}                &                        \multicolumn{1}{c}{\checkmark}&                        \multicolumn{1}{c}{\checkmark}&                     \multicolumn{1}{c}{\checkmark}\\
bird      & \multicolumn{1}{c}{\checkmark}               &                &                &                        &                        &                     \\
bus       & \multicolumn{1}{c}{\checkmark}               & \multicolumn{1}{c}{\checkmark}                 & \multicolumn{1}{c}{\checkmark}                &                        \multicolumn{1}{c}{\checkmark}&                        \multicolumn{1}{c}{\checkmark}&                     \multicolumn{1}{c}{\checkmark}\\
car       & \multicolumn{1}{c}{\checkmark}               &                &                &                        &                        &                     \\
 cat& & & & \multicolumn{1}{c}{}& \multicolumn{1}{c}{}&\multicolumn{1}{c}{}\\
cow       & \multicolumn{1}{c}{\checkmark}               &                &                &                        &                        &                     \\
 dog& & & & & \multicolumn{1}{c}{\checkmark}&\multicolumn{1}{c}{\checkmark}\\
horse     & \multicolumn{1}{c}{\checkmark}               & \multicolumn{1}{c}{\checkmark}                &                \multicolumn{1}{c}{\checkmark}&                        \multicolumn{1}{c}{}&                        \multicolumn{1}{c}{}&                     \multicolumn{1}{c}{}\\
motorbike & \multicolumn{1}{c}{\checkmark}               & \multicolumn{1}{c}{\checkmark}                &                \multicolumn{1}{c}{\checkmark}&                        \multicolumn{1}{c}{\checkmark}&                        \multicolumn{1}{c}{\checkmark}&                     \multicolumn{1}{c}{\checkmark}\\
person    & \multicolumn{1}{c}{\checkmark}               & \multicolumn{1}{c}{\checkmark}                &                \multicolumn{1}{c}{\checkmark}&                        \multicolumn{1}{c}{\checkmark}&                        \multicolumn{1}{c}{\checkmark}&                     \multicolumn{1}{c}{\checkmark}\\
sheep     & \multicolumn{1}{c}{\checkmark}               &                &                &                        &                        &                     \\
tvmonitor & \multicolumn{1}{c}{\checkmark}               &                &                &                        &                        &                     \\ \hline
\end{tabular}\end{table*}

\subsection{IMPLEMENTATION DETAILS}
The training and inference processes are conducted on an NVIDIA GeForce RTX 3090 GPU, with the following software versions: Python: 3.8.18, CUDA: 11.3, Pytorch: 1.10.0, Ubuntu: 20.04.5 LTS. In the KD process, we adhere almost all the parameters set of FGD for anchor-free one-stage models, with the exception of adjusting the initial learning rate to $1\times10^{-3}$ for ResNet series backbone networks and $1\times10^{-2}$ for MobileNet series backbone networks , set the maximum epoch to 100, the batch size to 16 and the number of CPU threads to 4. In the RD process, we noticed that the convergence procedure of the larger model will be extremely difficult if we set $\alpha=0$. Thus, the hyperparameters are set as $\alpha = 1$ and $\delta = 2$, which means the knowledge of the Tutor~1 model and tutor~2 model mainly comes from the student model, with all other parameters consistent with KD process.

In the weight substitution process, we employ linear subtraction and linear addition to update the original tutor model, with the hyperparameters $\varepsilon$ and $\eta$ both being fixed at 1. During the re-distillation process, which updates the edge-AI model based on the updated tutor model on user's actual scenario, we set the learning rate at $1\times10^{-4}$, the maximum number of epochs at 20, the batch size at 8, and the temperature for fine-tuning at 1.5.

\subsection{EVALUATION METRICS}
We evaluate the DiReDi framework from both the model performance and knowledge learning performance perspectives. In addition to average precision (AP), mean average precision (mAP), precision, and recall, we use the $F_{1}$ score to comprehensively assess the models' ability to predict positive and negative instances. The $F_{1}$ score is calculated using the following formula
\begin{equation}
F_{1}=\frac{2 \times \text { Precision } \times \text { Recall }}{\text { Precision }+ \text { Recall }}.
\end{equation}

% $L_{2}$ norm, can be defined as follows, is also used to compare the improvement in AP for different categories of objects before and after updating the Student Model. 
% \begin{equation}
% L_{2}=\sqrt{\sum_{i=1}^{n}\left(AP^{\text {Updated Student Model}{ }_{}}{ }_{i}-AP^{\text {Original Student Model }}{ }_{i}\right)^{2}} 
% \end{equation}

\section{PERFORMANCE}
In this section we examine the performance of different models in the DiReDi framework. The results are listed in Table~\ref{tab:table}. We will conduct two experiments with different subsets of data to examine the feasibility of gaining new knowledge and the consequence of forgetting knowledge of the process. 

\begin{table*}[]
\centering 
\caption{The Performance Metrics of different Models in the experiments}
\label{tab:table} 
\begin{tabular}{ccccc}
\hline
Model& mAP (\%) & Precision (\%) & Recall (\%) & F1 (\%) \\ \hline
Large& 77.8     & 84.9           & 71.5        & 77.6   \\
Tutor after Distill A with presumed data& 69.6     & 80.7           & 60.5        & 69.1    \\
Edge-AI after Distill B with presumed data& 55.3     & 67.9           & 50.3        & 57.8 \\ \hline
Experiment 1\\
Tutor 1 after ReDi B with presumed data& 53.7& 62.3& 49.1&54.9\\
Tutor 2 after ReDi A with presumed data and privacy data& 56.3& 64.0& 52.1& 57.4\\
Original Tutor with presumed data and privacy data & 57.0& 65.7& 50.4&57.0\\
Updated Tutor with presumed data and privacy data& 58.7& 70.6& 52.9& 60.5\\ 
Original edge-AI model& 43.8& 53.1& 41.8&46.8\\
Updated edge-AI Model (Training)& 53.9& 64.5& 50.5&56.7\\ 
Updated edge-AI Model (Distill C)& \textbf{59.9}& \textbf{70.6}& \textbf{55.3}& \textbf{62.0}\\\hline
Experiment 2 \\
Tutor 1 after ReDi B with presumed data& 57.3& 64.5& 51.8&57.5\\
Tutor 2 after ReDi A with presumed data and privacy data& 56.7& 64.3& 51.7& 57.3\\ 
Original Tutor with presumed data and privacy data& 55.8& 65.2& 49.0&55.9\\
Updated Tutor with presumed data and privacy data& 59.5& 67.9& 56.2& 61.5\\ 
Original edge-AI Model & 46.3& 56.3& 42.3&48.3\\
Updated edge-AI Model (Training)& \textbf{66.3}& 76.0& 57.0&65.2\\
Updated edge-AI Model (Distill C)& 66.2& \textbf{78.3}& \textbf{58.5}& \textbf{67.0}\\\hline
\end{tabular}
\end{table*}

\subsection{KNOWLEDGE DISTILLATION}
\begin{figure}[t]
    \centering
    \includegraphics[width=\linewidth]{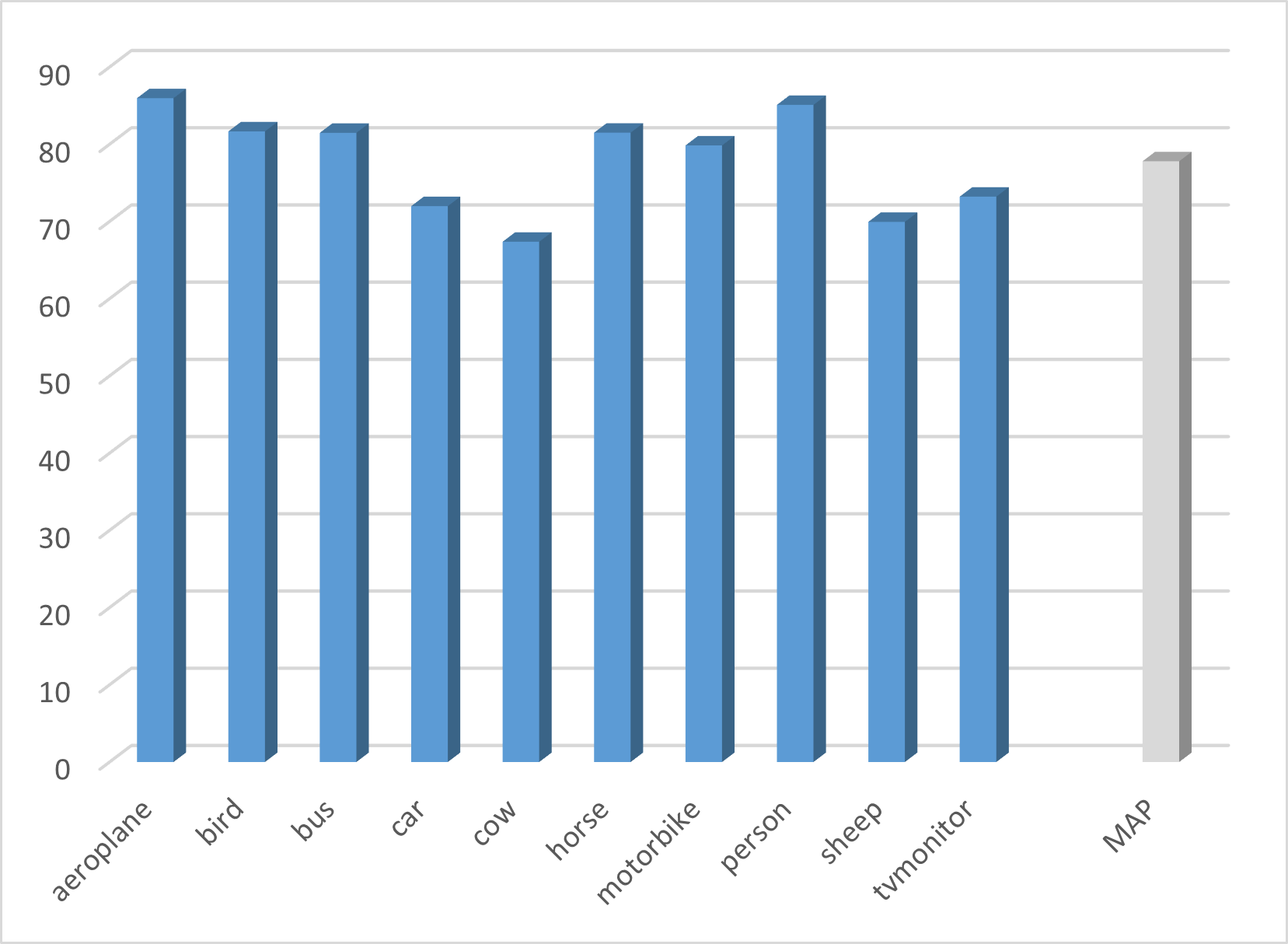}
    \caption{AP of the large Model on a dataset of 10 object categories.}
    \label{fig:2}
\end{figure}

\begin{figure}[t]
    \centering
    \includegraphics[width=\linewidth]{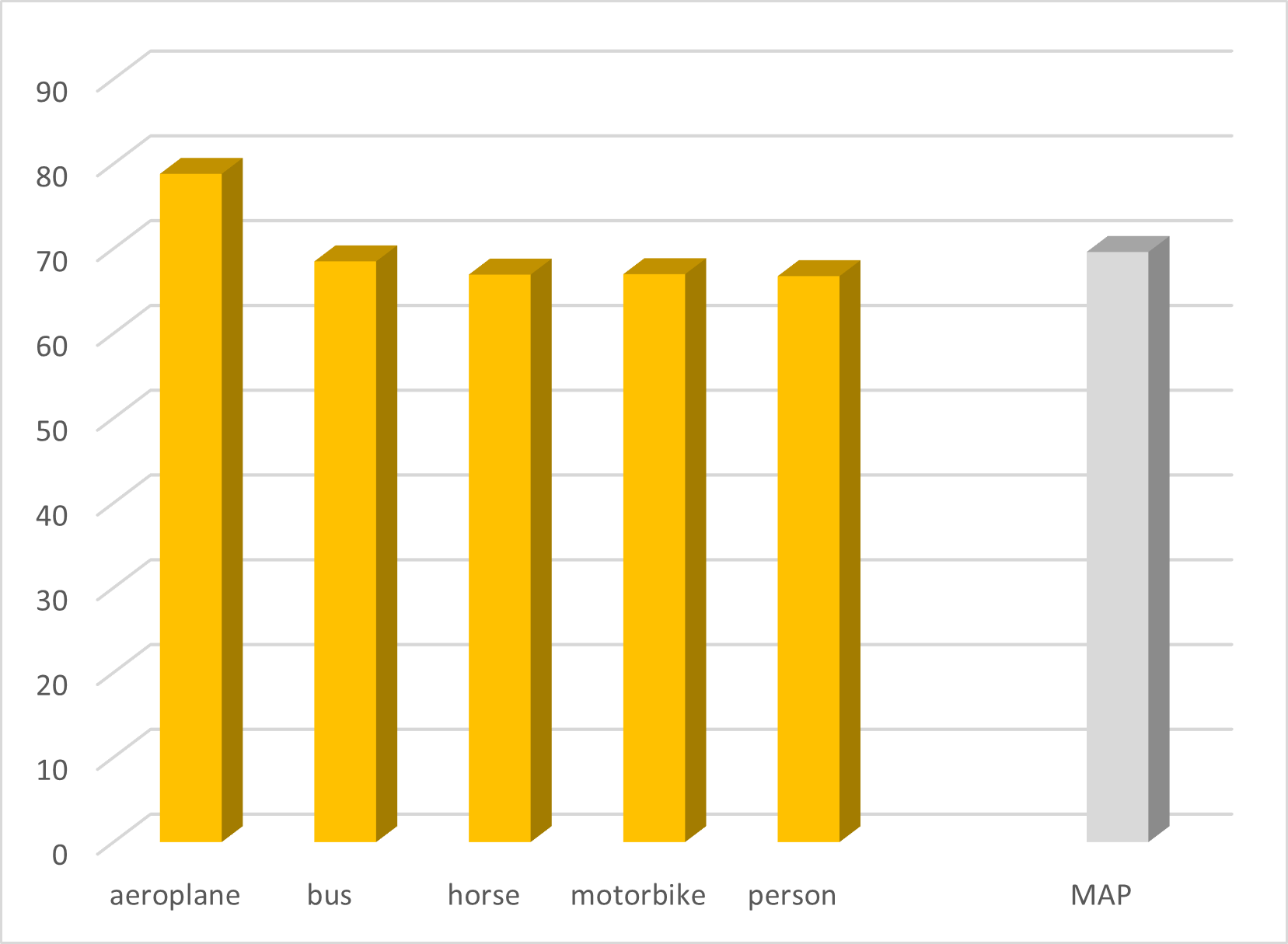}
    \caption{AP of the teacher model on the dataset of 5 object categories.}
    \label{fig:3}
\end{figure}

\begin{figure}[t]
    \centering
    \includegraphics[width=\linewidth]{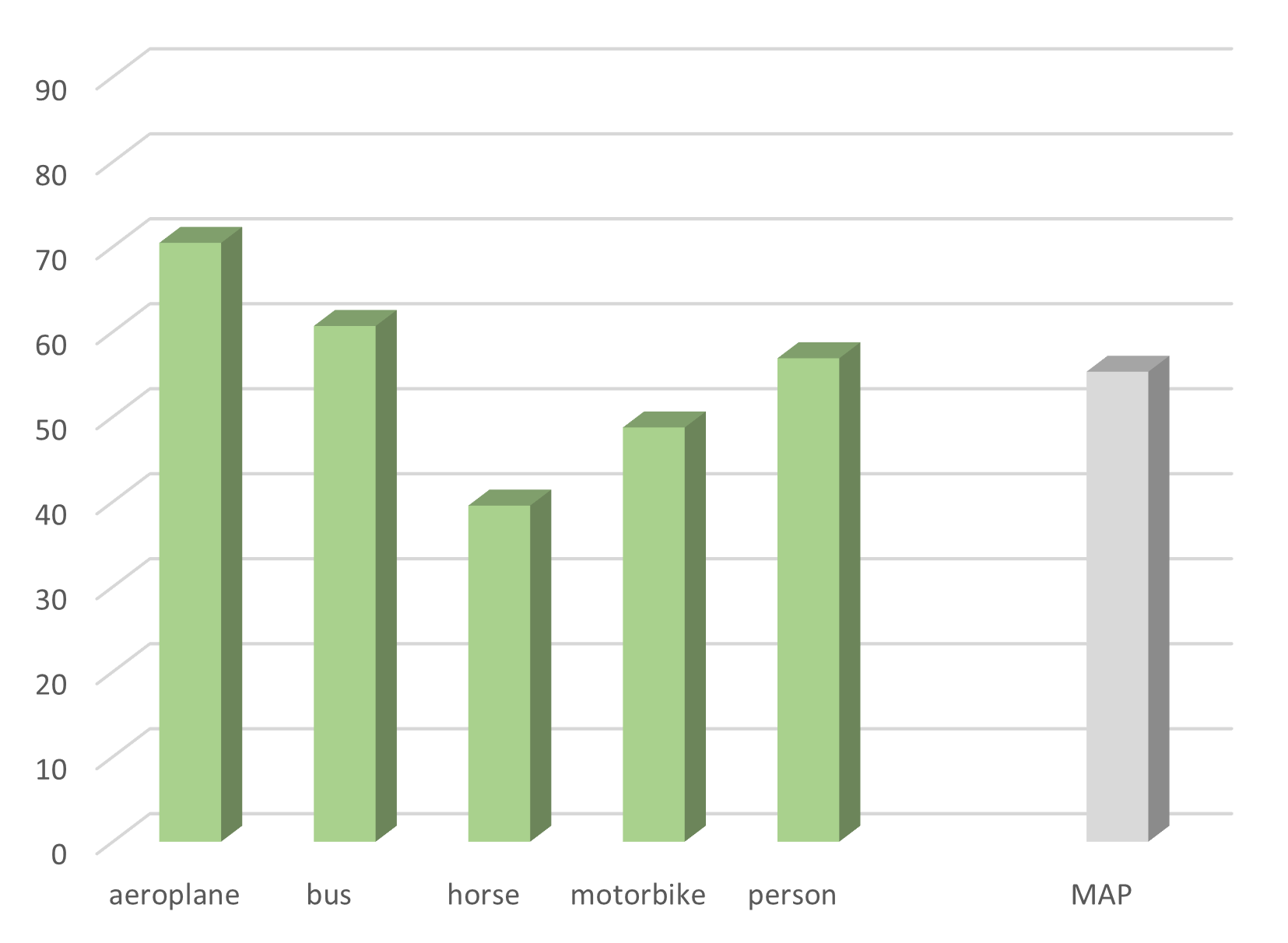}
    \caption{AP of the student model on the dataset of 5 object categories.}
    \label{fig:4}
\end{figure}

A ResNet101 as a large model is trained on 10 categories of object dataset, as shown in Table~\ref{tab:t1}. The specific performance metrics are shown in Fig. \ref{fig:2}.
The tutor Model is obtained through distillation from the large Model, using the presumed dataset that contains 5 object categories of targeted specific scenario. The specific performance metrics are shown in Fig. \ref{fig:3}. These categories are selected based on the assumption of user's application scenario. Note that this assumption might be different from what user will have in actual application scenario. After secondary distillation, the performance of the edge-AI Model, suitable for deployment in AIoT equipment in the manufacture domain, is shown in Fig.~\ref{fig:4}.

\subsection{EXPERIMENT 1, LEARNING NEW KNOWLEDGE}
\begin{figure}[t]
    \centering
    \includegraphics[width=\linewidth]{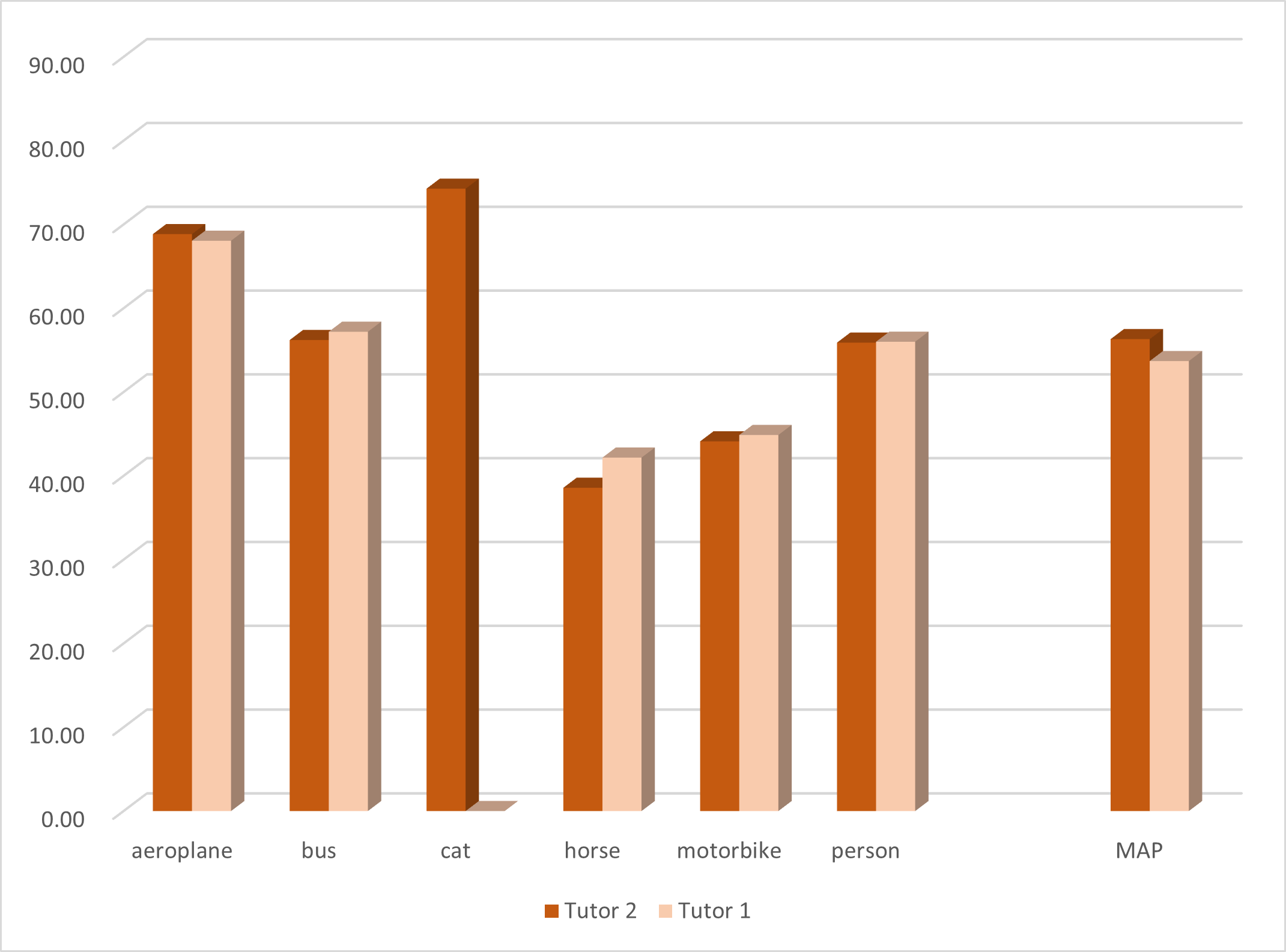}
    \caption{AP of both tutor 1 model and tutor 2 model in experiment~1.}
    \label{fig:5}
\end{figure}

Subsequently, Experiment 1 is conducted to verify the capability for learning new knowledge in the DiReDi framework. The models of both tutor 1 and tutor 2 are derived using the RD process, with the results shown in Fig. \ref{fig:5}. Note that the privacy data from the "cat" category is added while keeping the presumed data for the training of tutor 2 only. From this figure we can see that tutor 1 doesn't recognize "cat" category which is only included in Private data. That indicates the tutor 1 emulates the performance of the edge-AI model. Whereas, tutor~2 behaves as expected by the customer which can recognize "cat". This is contributed by adding the privacy data in the RD~A process.

\begin{figure}[t]
    \centering
    \includegraphics[width=\linewidth]{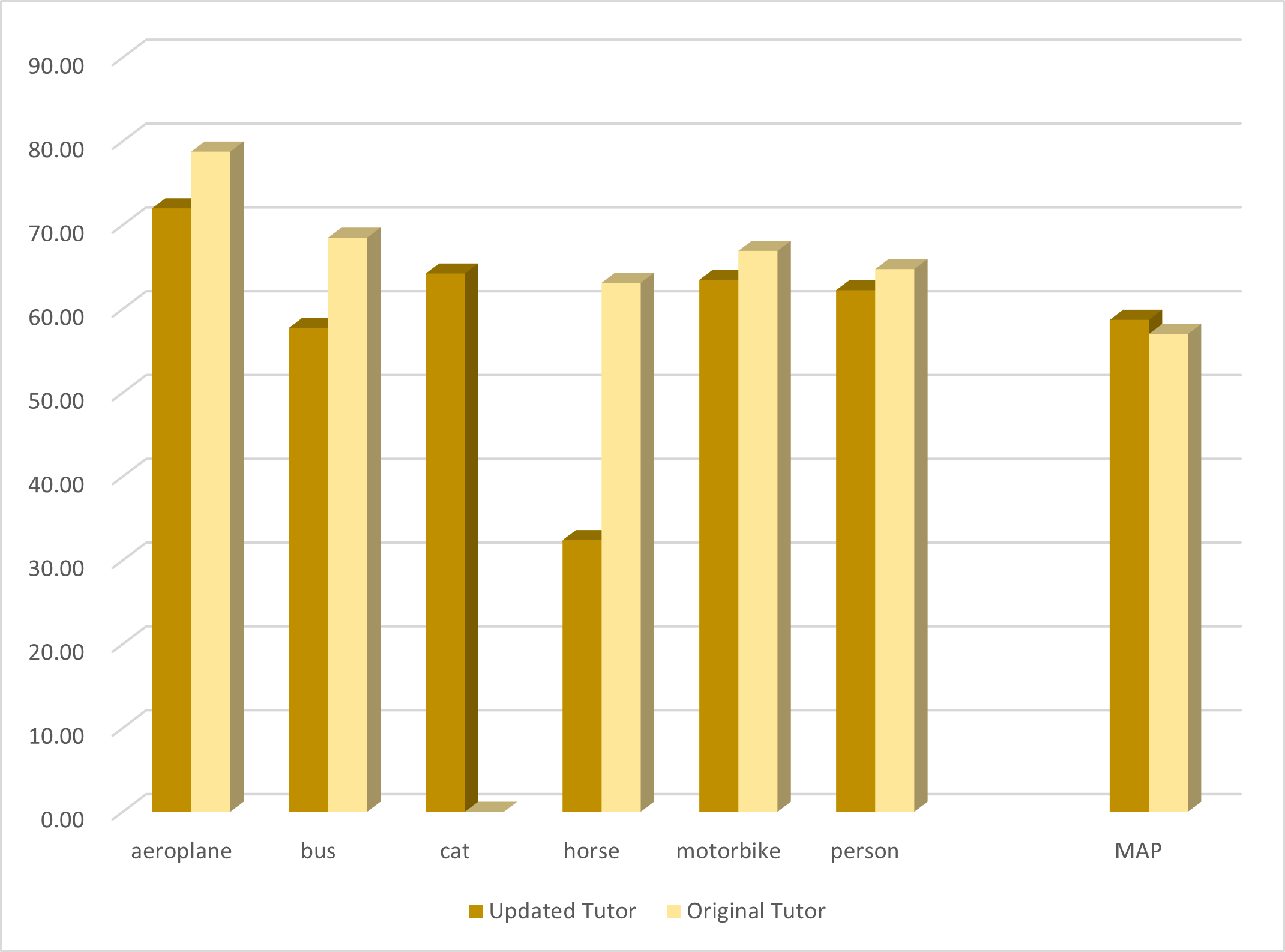}
    \caption{AP of the updated tutor model and original tutor model in experiment~1.}
    \label{fig:6}
\end{figure}

\begin{figure}[t]
    \centering
    \includegraphics[width=\linewidth]{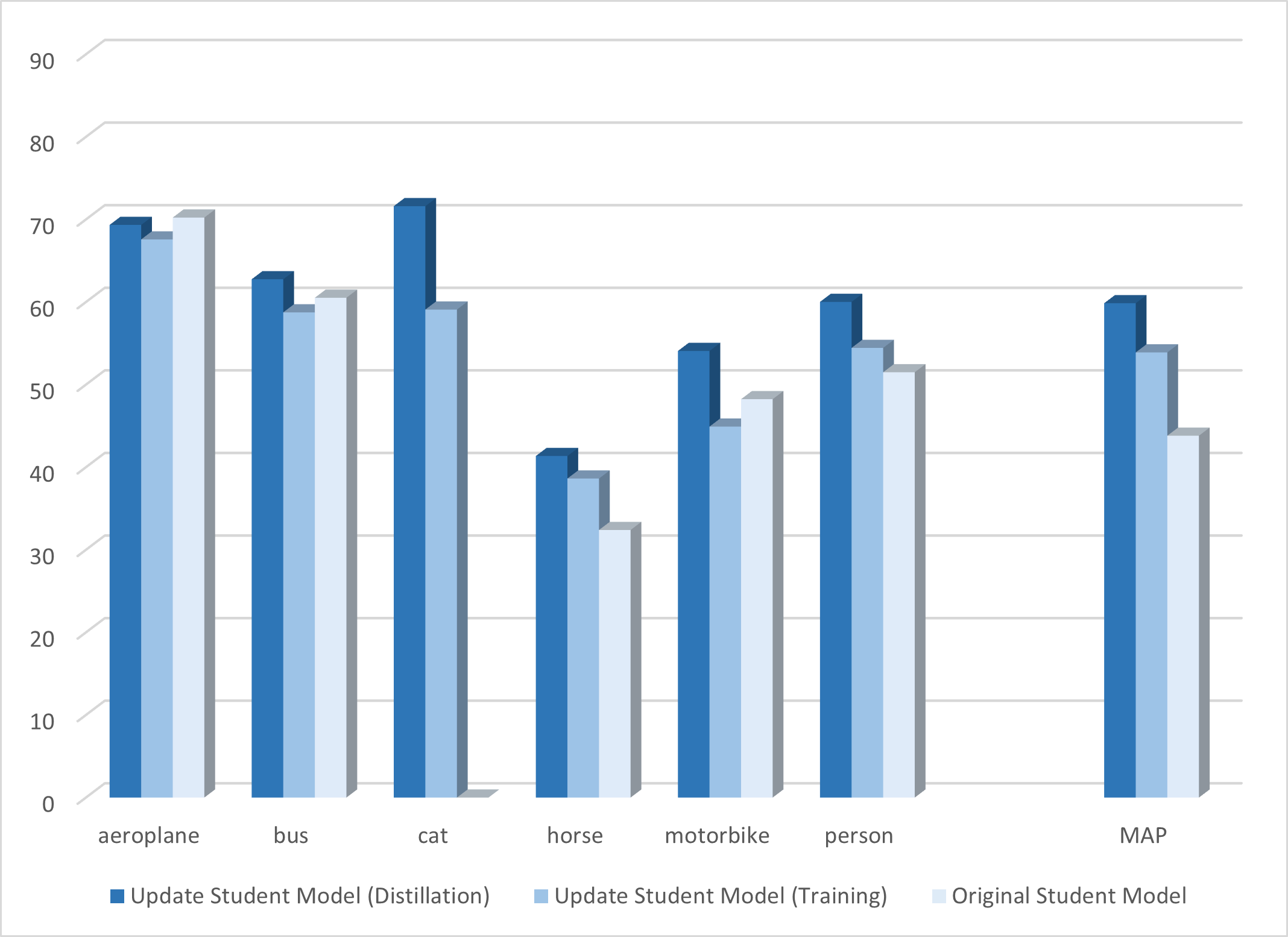}
    \caption{AP of the updated tutor model and the original tutor model in experiment~1.}
    \label{fig:7}
\end{figure}

After weight substitution, the performance of the tutor model based on the presumed dataset and private dataset are compared comprehensively as shown in Fig. \ref{fig:6}. Note that the performance of the original tutor is worse than the tutor after distillation A in that the original tutor is trained with only presumed data in the distillation A. Therefore, when there is new data different from the presumed data, the performance gets worse. Nevertheless, we can see that after the update from the RD, the data of new category is recognized and the performance improves. This process shows that we can obtain new knowledge from the user domain without privacy issue.
With the updated tutor carrying the new knowledge, we can apply the distillation B process to update the edge-AI model. Updating the tutor at manufacture side instead of updating the model on user's side allows the manufacture to check the performance of the tutor before using it to update the tutor model for customer. The updated tutor model can also be kept at the manufacture side for distilling edge-AI models to other customers. We also compare the performance of edge-AI model using distillation and direct training. Comparisons in Fig.~\ref{fig:7} indicate that the model obtained using this DiReDi method performs better than that from training directly on the local system.

\subsection{EXPERIMENT 2, LOSING KNOWLEDGE}

\begin{figure}[t]
    \centering
    \includegraphics[width=\linewidth]{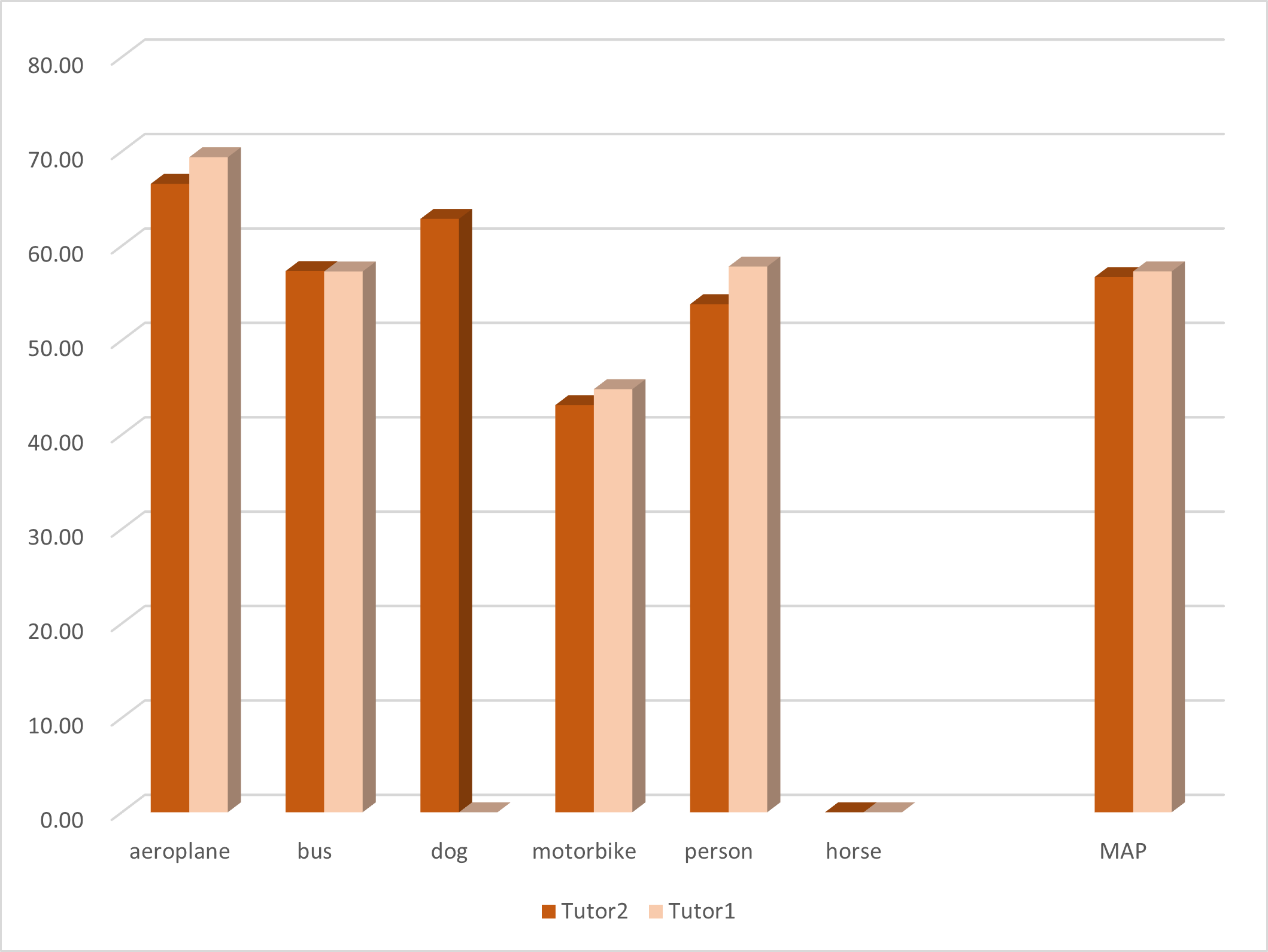}
    \caption{AP of the tutor~1 model and tutor~2 model in experiment 2.}
    \label{fig:9}
\end{figure}

Experiment 2 is conducted with data in Table~2 to demonstrate that the customer uses this AIoT devices in a new scenario which includes new data of "dog" category but doesn't have the usage of data from "horse" category. In the RD process, the data of "horse" is removed from the presumed data. The performance is shown in Fig.~\ref{fig:9}.

\begin{figure}[t]
    \centering
    \includegraphics[width=\linewidth]{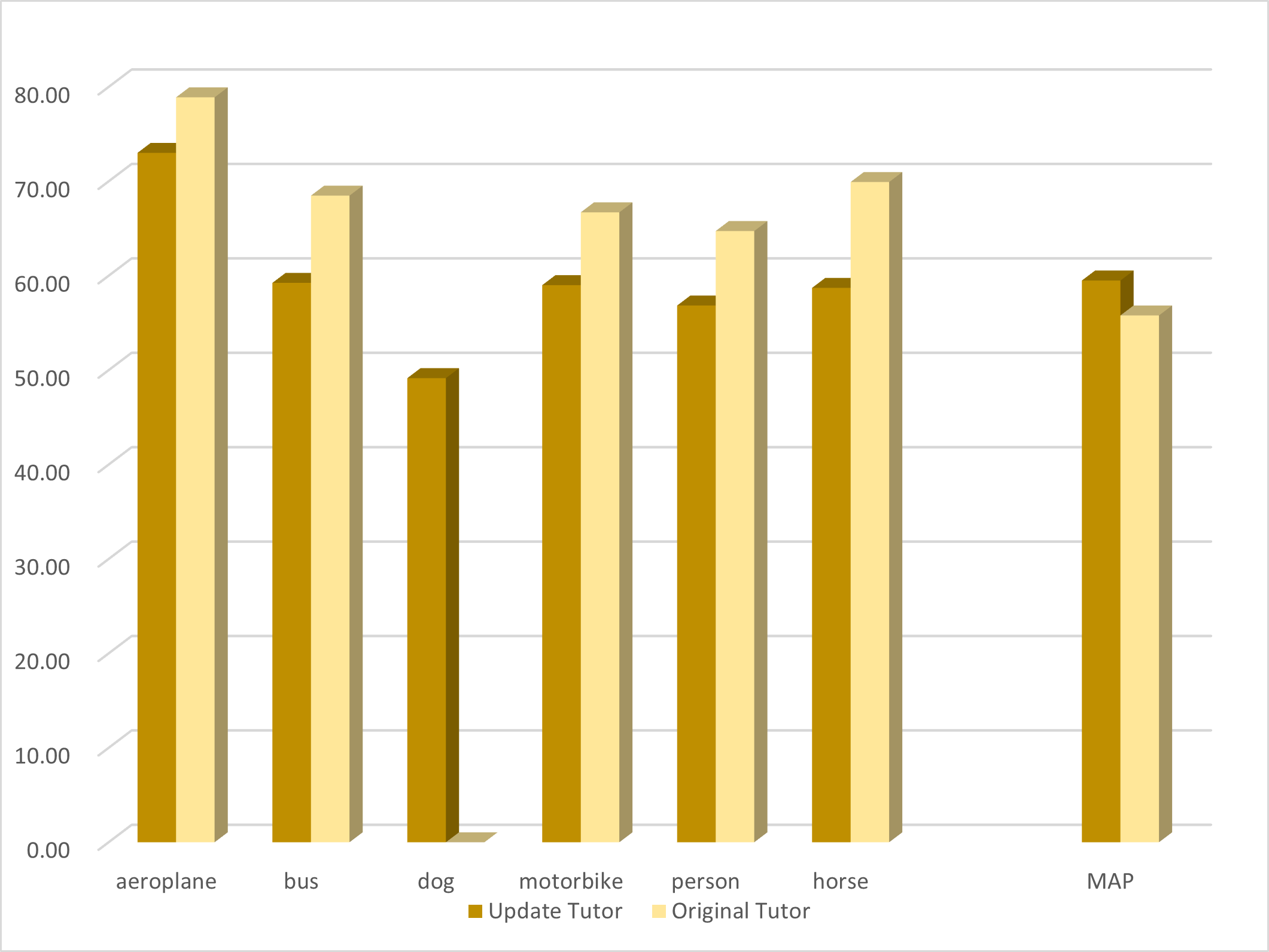}
    \caption{AP of the updated tutor model and the original tutor model in experiment~2.}
    \label{fig:10}
\end{figure}

After the weight substitution, the performance of the updated tutor model is compared with the original tutor model as shown in Fig.~\ref{fig:10}. The updated tutor learns new knowledge but the knowledge of "horse" reduced.

\begin{figure}[t]
    \centering
    \includegraphics[width=\linewidth]{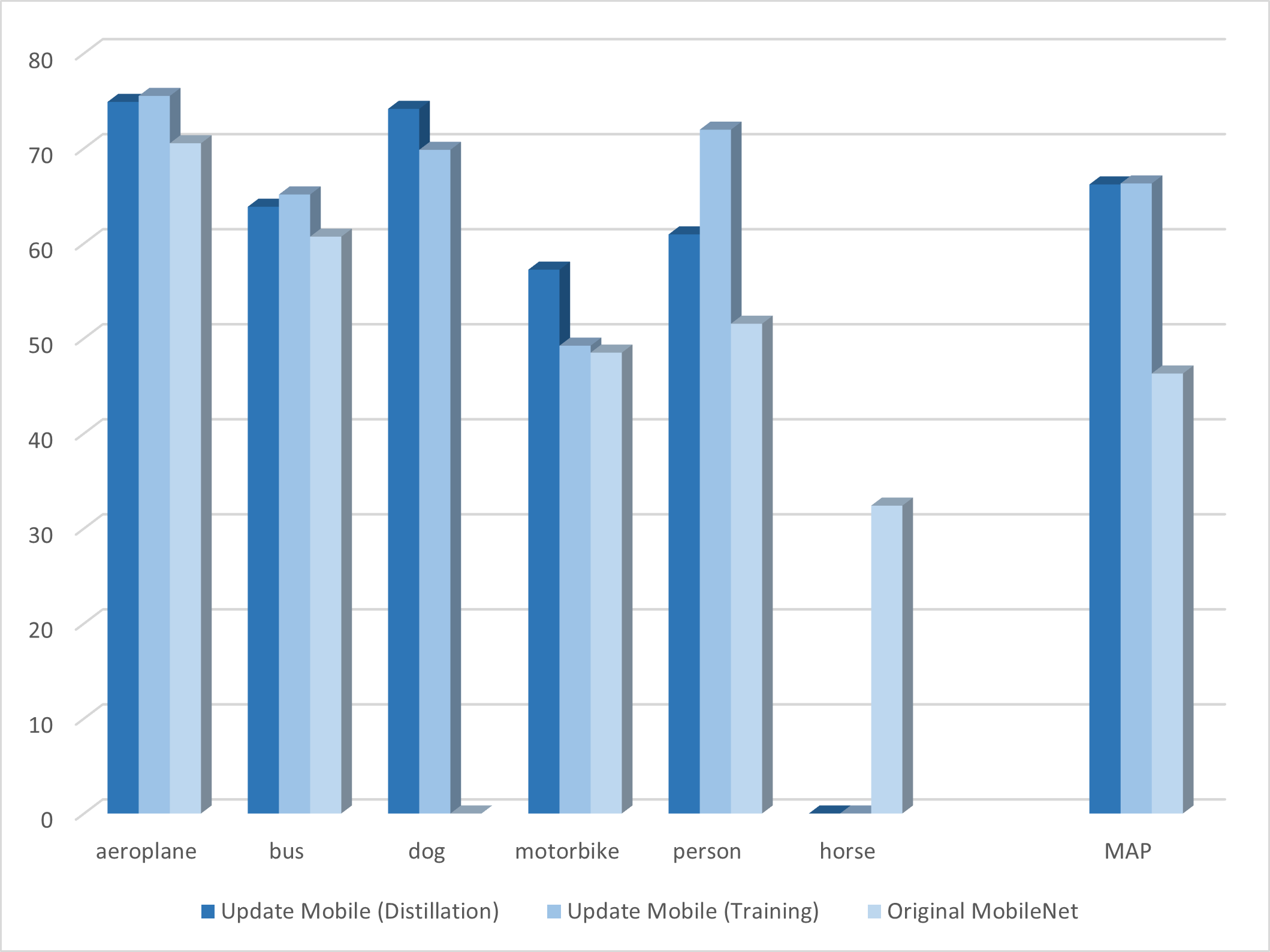}
    \caption{AP of the updated tutor model and the original tutor model in experiment~2.}
    \label{fig:12}
\end{figure}

Now, the AIoT company can check the knowledge of the updated tutor. If the knowledge of "horse" is critical, the manufacture can terminate this update process to avoid AI model failure for practical use. If this loss of knowledge is acceptable, the edge-AI model can be regenerated based on the updated tutor model. Again, the comparison of direct training and KD process is shown in Fig.~\ref{fig:12}. We can see that the edge-AI model has gained the new knowledge while started to losing the knowledge which might be useless from the customer perspective.

\subsection{SYSTEM COMPLEXITY}
It must be acknowledged that the system requires the deployment of greater computational power on the client to complete the update process, although our method consistently enhances the model’s F1 score without increasing model complexity. As shown in TABLE ~\ref{tab:t12}, taking the hardware and software equipment we used as an example, the GPU memory usage for directly training the edge-AI Model is 5.0~GB. For the DiReDi framework, the GPU memory usage for generating tutor~1 and tutor~2 through RD is 12.4~GB, while the GPU memory usage for the update distillation is 9.1~GB. This is closely related to the computational resource demands inherent in the distillation process itself.
\begin{table}[]
\centering 
\caption{Required Computational Resources in Experiment~2}
\label{tab:t12}
\begin{tabular}{ccc}
\hline
Method                 & Training Time & GPU Memory Usage \\ \hline
RevDistill A & About 19 h    & 12.4 GB          \\
RevDistill B & About 21 h    & 12.4             \\
Distill C    & About 4 h     & 9.1              \\ 
Directly Training      & About 4 h     & 5.0              \\ \hline
\end{tabular}
\end{table}

\section{CONCLUSIONS}
In this paper, we propose a framework to update the local model of AIoT. This framework features two RD processes. Through this process, the edge-AI model deployed on user's actual scenario can express the discrepancy between its internal knowledge and the knowledge provided by the actual scenario. This discrepancy is represented by the adjustment of parameter values from 2 tutor models that we used to emulate edge-AI model using presumed data and private data, respectively. By uploading this information to the manufacturer's server, the new information learned by the customer can be shared with the manufacturer. This process allows the manufacturer to validate the knowledge update before retraining the edge-AI model in the AIoT. 

We have studied the situation where the data from the user application scenario differs from the presumed data. We assumed that the data used to train the tutor model is still available, meaning the original tutor model retains the knowledge. In the distillation process, we used only a subset of this training data. In future work, we aim to extend DiReDi to handle situations where the data is entirely new, even to the tutor model.

Another direction is to apply DiReDi to wireless communication. A phone manufacturer might have trained a small model for mobile phones. It can be envisioned that the training data from a simulated wireless environment differs from the actual wireless environment. Further work is needed to define how to distill AI models for wireless communication and how to obtain the discrepancy in mobile phone performance in real-world wireless environments.
\bibliographystyle{IEEEtran}
\bibliography{ref}
\end{document}